\documentclass{speechpreprint}

\usepackage{caption}
\usepackage{subcaption}
\usepackage{graphicx}
\usepackage{enumitem}
\usepackage{amsmath}
\usepackage{booktabs}
\usepackage{hyperref}
\usepackage{cleveref}
\usepackage{amssymb}
\usepackage{tikz}
\usetikzlibrary{positioning, calc, arrows}

\title{An automated pipeline for standardised speech-unit annotation in spontaneous dialogue}

\author[affiliation={1}, orcid=0009-0008-5637-3574,correspondingauthor]{Hanlu}{He}
\author[affiliation={1}, orcid=0009-0002-6136-8207]{Harald Vilhelm}{Skat-Rørdam}
\author[affiliation={1},orcid=0000-0002-2222-0739]{Ingvi}{Örnólfsson}
\author[affiliation={1},orcid=0000-0003-3405-6291]{Ivana}{Konvalinka}

\address{
    $^1$ Department of Applied Mathematics and Computer Science, Technical University of Denmark
}

\email{hahea@dtu.dk, harsk@dtu.dk, rinor@dtu.dk, ivako@dtu.dk}


\usepackage{comment}

\begin{document}

\maketitle

\begin{abstract}

Quantifying conversational dynamics requires reliable identification of interactional units and their temporal boundaries, but speech activity alone does not distinguish conversational turns from listener feedback or within-turn pauses. We present an automated pipeline for extracting turns and backchannels from separate-channel recordings of spontaneous dyadic conversation, designed to provide a consistent first-pass annotation for subsequent human review. The pipeline combines voice activity detection, channel-energy filtering, temporal merging, automatic speech recognition, and context-based post-processing. We evaluated the pipeline on 99 ten-minute Danish conversations from 33 dyads using segment-level detection reliability and temporal boundary error. Conversations were recorded under both normal and asymmetric listening conditions. In the latter, speech-shaped noise was delivered to one participant through bone-conduction headphones. Overall detection reliability was $F_1=0.621$, with similar performance for turns ($F_1=0.624$) and backchannels ($F_1=0.618$). For successfully matched segments, median absolute onset and offset errors were 0.150 and 0.160\,s for turns and 0.130 and 0.180\,s for backchannels, respectively. Mean errors were substantially larger for turn boundaries, indicating a smaller number of large boundary mismatches. Performance did not differ significantly across the two experimental listening conditions. In a four-conversation case study, pipeline--human agreement was lower and more variable than human inter-annotator agreement and varied across parameter settings. These results support the pipeline as an automated first pass within a semi-automated annotation workflow, providing a consistent basis for more standardised and reproducible annotation of conversational dynamics.

\end{abstract}

\section{Introduction}

Spontaneous conversation is a highly coordinated form of social interaction in which speakers continuously manage when to speak, when to yield the floor, and how to respond - all while listening to another person speaking \cite{sacks1974simplest,stivers2009universals,levinson2016turn,meyer2023timing}. Conversational dynamics can be quantified through both the occurrence of interactional events, such as turns and backchannels, and their temporal organization, including turn duration, transition timing, pauses, and overlap. These measures characterize how interlocutors distribute and coordinate their contributions over time and have been associated with social connection, rapport, and interaction quality \cite{templeton2022fast,yokozuka2021relationship,west2025high}. They are also relevant to applied areas such as hearing and clinical research \cite{petersen2022effects,van2022communication}. Reliable identification of these interaction-level units from recordings is therefore an important methodological step in the analysis of natural conversation.

Extracting such units directly from conversational audio is challenging. While voice activity detection (VAD) can identify regions containing speech, speech activity does not map directly onto conversational structure. Pauses can occur within an ongoing turn, speaker transitions can occur with little or no silence, and genuine contributions from two speakers can overlap \cite{heldner2010pauses,stivers2009universals}. In separate-channel recordings, speech from one participant may also leak into the interlocutor's microphone, creating apparent speech activity on both channels \cite{ghosh2010robust}. An automated method must therefore distinguish recording artefacts and fragmented speech activity from interactionally meaningful speaker contributions. This distinction is particularly important because conversational turns are not defined by acoustic boundaries alone. Turn transitions are jointly organized and can depend on syntactic, lexical, prosodic, and temporal information \cite{sacks1974simplest,selting2000construction, levinson2016turn,gravano2011turn}. Consequently, a pause does not necessarily terminate a turn, and a new speech segment does not necessarily represent a new attempt to take the floor. Segmentation based only on speech activity can therefore produce acoustically valid regions that do not correspond to the units required for analysis of conversational dynamics.

Listener responses introduce an additional classification problem. Backchannels, for example, provide feedback to the current speaker without attempting to take the conversational floor \cite{kjellmer2009we,mereu2024backchannels}. They may occur during an ongoing turn and vary in lexical form, duration, and complexity \cite{gardner2001listeners,mereu2024backchannels}. Some consist of short conventional responses such as \textit{mhm} or \textit{yeah}, whereas
others contain multiple units and extend beyond the duration expected of a minimal response \cite{mereu2024backchannels}. Hence, backchannels cannot be identified using duration or a fixed lexical dictionary alone, as their classification depends on conversation context.

How these units are operationalized directly affects downstream measures of conversational dynamics. Correct classification determines which events contribute to measures such as turn and backchannel frequency, while temporal boundary detection affects measures of turn duration, transition timing, pauses, and overlap. Reliable analysis therefore requires both accurate identification of conversational units and sufficiently precise temporal segmentation. Differences in these operational criteria can also limit comparison across studies. Consistent segmentation criteria are therefore important for both annotation reliability and the reproducibility of quantitative conversation research \cite{heldner2010pauses,meyer2023timing,mereu2024backchannels}.

These requirements create a gap between the output of general speech-processing tools and the interaction-level labels needed for conversation analysis. VAD estimates when speech is present, while automatic speech recognition (ASR) estimates what was said. Neither alone determines whether detected speech is a continuation of an existing turn, a new floor-holding contribution, or listener feedback. Recent work has begun to address this problem at the transcript level. NaturalTurn \cite{cooney2025naturalturn} is a deterministic post-processing method that transforms speaker-specific ASR output into more naturalistic conversational turns. It merges successive utterances belonging to the same floor-holding contribution and separates listener speech occurring within these turns into non-primary contributions, including backchannels and longer secondary turns. Backchannels are identified using configurable lexical cues and simple rules based on the content and length of the listener contribution. In this way, NaturalTurn enables the reconstruction of conversational structure from already transcribed speech without requiring annotated training data. 

The problem that our pipeline aims to solve begins at an earlier stage of processing. In separate-channel recordings of face-to-face conversation, the speaker-specific speech segments from which such transcripts are constructed must first be identified. Cross-channel leakage can produce apparent speech activity on both channels, while natural within-turn pauses can fragment a single conversational contribution into multiple detected speech segments \cite{ghosh2010robust,wang2024cross,dourado2026gamma}. Deriving turns and backchannels from the audio therefore requires cross-channel leakage to be distinguished from
target-speaker activity and fragmented speech segments to be merged before
conversational units can be classified. Unlike NaturalTurn, which takes speaker-specific word-level ASR output as its input, our pipeline derives speaker-specific temporal units directly from the audio, with transcription used subsequently to help classify these units.

The present study therefore develops an automated audio-to-annotation pipeline for extracting turns and backchannels from separate-channel recordings of spontaneous dyadic conversation, where participants converse freely without a prescribed topic or task. The pipeline combines VAD, channel-energy filtering, temporal merging, automatic transcription, and conversational-unit classification. Backchannels are identified from the transcribed speech and its conversational context rather than from a predefined list of backchannel expressions or simple thresholding. The resulting annotations are intended as a consistent first pass for subsequent human review.

For the present analysis, a \emph{turn} is defined as a continuous contribution by one participant while holding the conversational floor. Within-turn pauses, hesitations, and filled pauses do not necessarily terminate the turn. A \emph{backchannel} is defined as a listener contribution that provides feedback to the current speaker without attempting to take the conversational floor \cite{duncan1974signalling,mereu2024backchannels}.

We evaluate the pipeline on 99 ten-minute face-to-face Danish conversations from 33 dyads. The evaluation addresses two aspects required for quantifying conversational dynamics: whether turns and backchannels are correctly identified, and how accurately their temporal boundaries are located. We additionally examine performance across listening conditions and parameter settings. A subset of four conversations independently annotated by a second human annotator provides an initial benchmark against human inter-annotator agreement.

\section{Methods}

\subsection{Conversation unit definitions}\label{sec:def}

Two conversational units were defined, in order to guide the human annotators:

\begin{itemize}
    \item \textbf{Turn}: A continuous stretch of speech produced by a single participant before yielding the conversational floor. Minor pauses, hesitations, and filled pauses within the same floor-holding contribution do not constitute separate turns.

    \item \textbf{Backchannel}: A listener-produced utterance that provides feedback relevant to the ongoing turn without attempting to take the conversational floor. Backchannels may be lexical (e.g., \textit{yes}, \textit{okay}) or non-lexical (e.g., \textit{uh-huh}, \textit{mm}), and may vary in form, duration, and function \cite{mereu2024backchannels}.
\end{itemize}

Two additional terms were used for the technical implementation of the annotation pipeline:

\begin{itemize}
    \item \textbf{Crosstalk}: Acoustic leakage in which speech produced by one participant is captured on the interlocutor's microphone channel. Crosstalk is treated as a recording artefact and is distinguished from genuine conversational overlap, in which both participants produce speech simultaneously.

    \item \textbf{Overlapped turn}: An algorithmic label assigned to a candidate turn that is fully temporally contained within a turn produced by the interlocutor. The label describes the temporal relationship between the two segments and does not, by itself, imply successful or attempted floor-taking.
\end{itemize}

\subsection{Human reference annotation}

Human reference annotations were produced manually using the conversational-unit definitions described in Section ~\ref{sec:def}. One annotator annotated all 99 conversations and these annotations served as the human reference for the main pipeline evaluation.

The two participant audio channels were imported separately into ELAN. During annotation, the annotator could inspect both waveforms and switch between the two audio channels, allowing contributions from one participant to be interpreted in relation to the speech of the interlocutor. Each conversational unit was manually labelled as a turn or backchannel, and its onset and offset were marked on the corresponding participant channel by listening to the audio while inspecting the waveform.

To investigate inter-annotator agreement, a second annotator independently annotated four conversations from distinct dyads using the same definitions and annotation procedure. These annotations were compared with the corresponding annotations from the primary annotator using the same segment-matching criteria subsequently used for pipeline – human evaluation.

\subsection{Pipeline architecture}

The automated annotation pipeline transforms separate audio channels from each participant into a structured set of speech events labelled as turns or backchannels. The pipeline consists of seven stages, as illustrated in Fig.~\ref{fig:pipeline}. The stages combine speech activity detection, channel-energy filtering, automatic transcription, and rule-based post-processing to produce an initial annotation that can subsequently be reviewed by a human annotator.

In \textbf{Stage I}, mild audio preprocessing is applied separately to each participant's original audio channel. A 5th-order Butterworth high-pass filter with a cut-off frequency of 60\,Hz is applied using the \texttt{scipy} package \cite{2020SciPy-NMeth}. This removes low-frequency signal components while otherwise limiting modification of the audio prior to voice activity detection and channel-energy comparison.

In \textbf{Stage II}, speech activity is detected separately in each participant's audio channel using rVAD \cite{tan2020rvad}. rVAD was selected because it is an unsupervised and computationally lightweight VAD method designed for robust speech activity detection across variable acoustic conditions, without requiring task-specific training data. This makes it suitable as an initial speech-detection stage in the present pipeline, where subsequent processing further resolves cross-channel leakage and conversational structure.

Speech activity is estimated on 25-ms frames with a 10-ms step size. The rVAD decision threshold, denoted $\tau_{\mathrm{VAD}}$, controls the criterion used by the algorithm to classify frames as speech or non-speech. The pipeline uses the default $\tau_{\mathrm{VAD}} = 0.4$. The sensitivity of pipeline performance to this parameter is examined separately in the parameter analysis.

In \textbf{Stage III}, an energy-based filter is applied to reduce crosstalk between the two participant channels. For each speech segment $k$ identified by rVAD, the root-mean-square (RMS) amplitude is calculated as

\begin{equation}
    \mathrm{RMS}_k =
    \sqrt{\frac{1}{N_k}\sum_{n=1}^{N_k}x_k[n]^2},
\end{equation}

where $x_k[n]$ is the amplitude of sample $n$ in segment $k$, and $N_k$ is the number of samples in that segment. RMS amplitude is converted to decibels as

\begin{equation}
    E_{\mathrm{dB},k}
    =
    20\log_{10}\left(\mathrm{RMS}_k+\epsilon\right),
\end{equation}

where $\epsilon=10^{-8}$ prevents taking the logarithm of zero.

Because absolute recording levels may differ between participants, the filtering threshold is defined separately for each speaker channel $c$ relative to the maximum segment energy observed on that channel:

\begin{equation}
    \theta_c = E_{\max,c} - \tau_{\mathrm{dB}},
\end{equation}

where $E_{\max,c}$ is the maximum segment energy observed for channel $c$. The value $\tau_{\mathrm{dB}}=50$\,dB was selected during preliminary inspection of recordings not included in the evaluation, as a conservative threshold that retained low-amplitude target speech while excluding low-energy cross-channel leakage. Speech segments for which $E_{\mathrm{dB},k}<\theta_c$ are removed from that channel before subsequent merging.

In \textbf{Stage IV}, VAD segments are merged into longer candidate conversational units using a 3-s forward-looking window. Same-speaker segments separated by $\leq0.5$\,s are merged directly. Segments separated by longer gaps may also be merged when their combined speech duration equals or exceeds the intervening gap, preventing longer contributions from being fragmented by pauses in speech activity. At this stage, merged segments are capped at 60\,s for computational efficiency during subsequent Whisper transcription. This is a processing constraint rather than a maximum conversational-unit duration, and no corresponding duration limit is imposed on the final turns
or backchannels.

Segments shorter than 1.0\,s are treated as \textit{short utterances} for the purposes of this merging procedure. When a short utterance from an interlocutor occurs between two segments from the same speaker, the algorithm can bridge across it and continue the original speaker's candidate unit. The intervening utterance is nevertheless retained as a separate candidate event, allowing potential backchannels and overlapping speech to be classified in subsequent stages.

The 0.5-s gap threshold is a theory-informed criterion rather than a definition of a turn boundary. Danish has comparatively long response offsets in question--response sequences \cite{stivers2009universals}, while backchannels can consist of multiple units and extend beyond 500\,ms \cite{mereu2024backchannels}. The merging procedure therefore allows temporal flexibility while reducing fragmentation introduced by frame-level speech activity detection.

In \textbf{Stage V}, stronger audio preprocessing is applied to the original audio channels prior to automatic transcription, to provide a more consistent input signal for ASR. The preprocessing consists of the 60-Hz high-pass filter described in Stage I, followed by stationary spectral-gating noise reduction using the \texttt{noisereduce} package \cite{sainburg2020finding}. The resulting signal is then loudness-normalized to $-23$\,LUFS following EBU R~128 \cite{ebu_loudness_2023}. Loudness normalization is skipped when the signal is already within $\pm3$\,LU of the target, and the applied gain is capped at $+20$\,dB to avoid excessive amplification. Finally, signal amplitude is limited to $\pm0.95$ (approximately $-0.45$\,dBFS) using \texttt{numpy.clip} \cite{harris2020array}. The processed audio is used only for the subsequent transcription stage. Speech-event boundaries remain those obtained from the previous stages.

In \textbf{Stage VI}, the preprocessed speech segments are transcribed using the multilingual \texttt{Whisper-large-v3} ASR model \cite{radford2022whisper}. In the pipeline, the transcription serves as an intermediate representation for the initial classification of candidate segments rather than as the final annotation output.

For each transcribed segment $k$, word-level Shannon entropy is calculated from the empirical distribution of word tokens:

\begin{equation}
    H_k = - \sum_{w \in V_k} p_k(w)\log_2 p_k(w),
    \label{eq:entropy}
\end{equation}

where $V_k$ is the set of unique word tokens in the transcription of segment $k$, and

\begin{equation}
    p_k(w) = \frac{c_k(w)}{M_k},
\end{equation}

where $c_k(w)$ is the number of occurrences of token $w$ in segment $k$ and $M_k$ is the total number of word tokens in that segment. Prior to entropy calculation, transcriptions are converted to lowercase and tokenized at the word level. Empty transcriptions and transcriptions containing a single token are assigned an entropy of $0$. This measure provides a simple proxy for lexical diversity within a segment: short or repetitive transcriptions tend to produce lower entropy, whereas segments containing a broader distribution of word types tend to produce higher entropy.

Segments with $H_k < \tau_H$ are initially classified as backchannels, whereas segments with $H_k \geq \tau_H$ are initially classified as turns. The pipeline default uses $\tau_H=1.5$; the sensitivity of pipeline performance to this parameter is examined separately in the parameter analysis section. These labels are preliminary and are subsequently re-evaluated using the criteria described in Stage VII.

In \textbf{Stage VII}, the preliminary classifications are refined using the temporal context of the two speakers. A segment initially classified as a turn is relabelled as an \textit{overlapped turn} when it is fully contained within a turn produced by the interlocutor.

The algorithm then performs a final context-based merging step. Two turns from the same speaker may be merged when the second begins within 3.0\,s of the current turn. Merging is allowed when no intervening segment is present, or when the intervening segments consist only of backchannels or overlapped turns with durations of at most 1.0\,s. These intervening segments are retained as separate annotations rather than absorbed into the merged turn.

Before evaluation, the pipeline output and human reference annotations were both subjected to the same context-based normalization. Same-speaker turns were merged unless they were separated by either a backchannel from that speaker or a turn from the other interlocutor. Backchannels that did not overlap with an interlocutor contribution were relabelled as turns, while turns fully contained within an interlocutor turn were relabelled as backchannels. These merging and relabelling rules were applied repeatedly until no further changes occurred, up to a maximum of 20 iterations.

Experiments were conducted using GPUs with at least 16\,GB of VRAM. Whisper-large-v3 inference was performed on the GPU, whereas rVAD processing was CPU-based. Processing a 10-min conversation through the complete pipeline required approximately 2 min on average, i.e.,  approximately $5\times$ real-time processing.

\begin{center}
    \vspace{0.01cm} 
    \captionsetup{type=figure} 
    \resizebox{\columnwidth}{!}{
    \begin{tikzpicture}[
        node distance=0.4cm and 0.8cm,
        auto,
        file/.style={draw, fill=gray!10, rectangle, minimum width=2.2em, font=\tiny, align=center},
        block/.style={rectangle, draw, fill=blue!10, text width=11em, text centered, rounded corners, font=\tiny},
        pre/.style={rectangle, draw, fill=yellow!15, text width=11em, text centered, rounded corners, font=\tiny},
        process/.style={rectangle, draw, fill=green!5, text width=13em, text centered, rounded corners, font=\tiny},
        line/.style={draw, -stealth, thick},
    ]

        \node[font=\tiny] (inputLabel) {Input: Separate-channel Audio Recordings};

        \coordinate[below=0.4cm of inputLabel] (fileCenter);
        \node[file] (fileA2) at (fileCenter) {Ch 2\\($P_2$)};
        \node[left=0.15cm of fileA2] (dotsA) {\dots};
        \node[file, left=0.15cm of dotsA] (fileA1) {Ch 1\\($P_1$)};
        \node[file, right=0.4cm of fileA2] (fileAn) {Ch N\\($P_n$)};

        \node[pre, below=0.5cm of fileA2] (premild) {Stage 1: Audio Preprocessing (Mild)};

        \draw[line] (fileA1.south) -- ++(0,-0.2) -| (premild.north);
        \draw[line] (fileA2.south) -- (premild.north);
        \draw[line] (fileAn.south) -- ++(0,-0.2) -| (premild.north);

        \node[block, below=0.3cm of premild] (vad) {Stage 2: Voice Activity Detection (VAD)};
        \draw[line] (premild) -- (vad);

        \node[process, below=0.3cm of vad] (viz) {
            \textbf{Segmented Speaker Recordings} \\ 
            \begin{tikzpicture}[xscale=0.3, yscale=0.2]
                \draw[->] (0,0) -- (8.5,0);
                \filldraw[fill=blue!30, draw=blue!80] (0.5, 2.2) rectangle (2.5, 2.8);
                \filldraw[fill=blue!30, draw=blue!80] (3.5, 2.2) rectangle (6.0, 2.8);
                \node[left, font=\tiny] at (0,2.5) {$P_1$};
                \filldraw[fill=red!30, draw=red!80] (2.0, 1.2) rectangle (3.0, 1.8);
                \filldraw[fill=red!30, draw=red!80] (6.5, 1.2) rectangle (7.5, 1.8);
                \node[left, font=\tiny] at (0,1.5) {$P_2$};
                \filldraw[fill=green!30, draw=green!80] (1.0, 0.2) rectangle (1.5, 0.8);
                \filldraw[fill=green!30, draw=green!80] (4.0, 0.2) rectangle (5.5, 0.8);
                \node[left, font=\tiny] at (0,0.5) {$P_n$};
            \end{tikzpicture}
        };
        \draw[line] (vad) -- (viz);

        \node[block, below=0.3cm of viz] (energy) {Stage 3: Energy Filtering};
        \node[block, below=0.2cm of energy] (merge) {Stage 4: Initial Merging};
        \node[pre, below=0.2cm of merge] (prestrong) {Stage 5: Audio Preprocessing (Strong)};
        \node[block, below=0.2cm of prestrong] (asr) {Stage 6: Transcription};
        \node[block, below=0.2cm of asr] (class) {Stage 7: Context-based Classification and Merging};

        \draw[line] (viz) -- (energy);
        \draw[line] (energy) -- (merge);
        \draw[line] (merge) -- (prestrong);
        \draw[line] (prestrong) -- (asr);
        \draw[line] (asr) -- (class);

        \node[below=0.4cm of class, draw, fill=orange!5, rounded corners, align=left, font=\tiny] (out) {
            \textbf{Segment labels}\\
            \begin{tabular}{@{}llllll@{}}
            \textbf{Segment} & \textbf{Spk} & \textbf{Start} & \textbf{End} & \textbf{Type} & \textbf{Text} \\ \hline
            $S_{est,1}$ & P1 & 3.71 & 8.03 & Turn & Hej, jeg... \\
            $S_{est,2}$ & P2 & 12.5 & 17.9 & Turn & Om et par... \\
            $S_{est,3}$ & P1 & 49.5 & 49.6 & BC & Ja. \\
            \end{tabular}
        };
        \draw[line] (class) -- (out);
    \end{tikzpicture}}
    
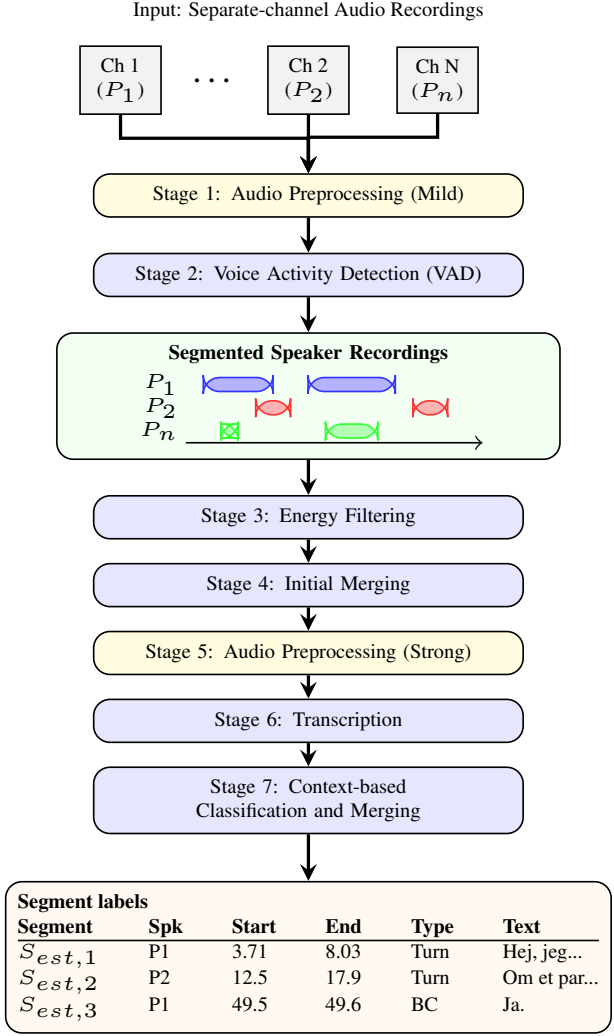
\captionof{figure}{Architecture of the automated pipeline for processing
separate-channel conversation recordings.}
    \label{fig:pipeline}
    \vspace{0.2cm}
\end{center}
 
\subsection{Dataset and recording protocol}

The dataset was collected as part of a larger multimodal study of spontaneous dialog between same-gender dyads. Participants were native Danish speakers between 18 and 35 years of age, with normal hearing and no reported communication difficulties, and were unacquainted prior to the experiment. The maximum age difference within a dyad was four years. Each dyad completed three 10-min face-to-face conversations in Danish, resulting in approximately 30 min of conversational speech per dyad. Participants were not assigned a topic or conversational task and were instructed to converse freely. Speech was recorded separately for each participant using head-worn directional microphones (DPA 4188) at 48~kHz. 

The three conversations were conducted in different listening conditions. In the noise-free condition, neither participant received experimental auditory interference. The two other conversations were conducted in an asymmetric-noise condition. In these, speech-shaped noise was delivered to only one participant through bone-conduction headphones, while the interlocutor conversed without auditory interference. The order of conditions was counterbalanced across dyads. Participants were informed that noise could occur during the experiment but were not told in advance when or to whom it would be presented, and were instructed not to disclose the manipulation during the conversation.

The level of the noise was calibrated using masked detection thresholds obtained in an independent sample of 10 participants aged 18--35 years. The level matched that which was required to achieve 70\% detection of a spoken digit presented at 45~dBA. Importantly, the noise from the bone-conduction headphones did not interfere with the recorded audio signal. Thus, the asymmetric noise condition represents variation in the listening conditions under which speech was produced, rather than variation in background noise within the recordings themselves.

The present analysis includes 99 conversations from 33 dyads, corresponding to approximately 16.5 hours of conversational audio. Four conversations from distinct dyads were additionally annotated by a second human annotator for evaluation of inter-annotator agreement.

\subsection{Evaluation Metrics}\label{sec:eval}

\subsubsection{Detection reliability}

Detection reliability was evaluated by matching automatically estimated segments to the human reference annotations using temporal \textbf{intersection over union (IoU)}. For a reference segment $S_{\mathrm{ref},i}$ and an estimated segment $S_{\mathrm{est},j}$, IoU was defined as

\begin{equation}
    \mathrm{IoU}(i,j) =
    \frac{
        \operatorname{dur}
        \left(S_{\mathrm{ref},i} \cap S_{\mathrm{est},j}\right)
    }{
        \operatorname{dur}
        \left(S_{\mathrm{ref},i} \cup S_{\mathrm{est},j}\right)
    }.
    \label{eq:iou}
\end{equation}

\begin{quote}
    \textit{Example: if a reference turn extends from 10.0 to 12.0\,s and the corresponding estimated turn extends from 10.5 to 12.5\,s, their intersection is 1.5\,s and their union is 2.5\,s, giving an IoU of $1.5/2.5=0.60$. Similarly, for a short backchannel extending from 10.0 to 10.5\,s in the reference annotation and from 10.4 to 10.9\,s in the pipeline output, the intersection is 0.1\,s and the union is 0.9\,s, giving an IoU of approximately 0.11.}
\end{quote}

Candidate matches were restricted to segments assigned to the same speaker and the same conversational-unit type. A reference and estimate segment were considered a match when their IoU was at least 0.50 for turns or 0.10 for backchannels. A lower threshold was used for  backchannels because their shorter durations make IoU more sensitive to small absolute differences in onset and offset timing.

When multiple candidate segments satisfied the overlap criterion, competing matches were resolved by retaining the candidate with the highest IoU, such that each segment contributed to at most one match. Matched reference segments were counted as true positives (TP), unmatched reference segments as false negatives (FN), and unmatched estimated segments as false positives (FP). Precision, recall, and $F_1$ scores were calculated from these classifications for turns and backchannels separately. An overall $F_1$ score was additionally calculated across turns and backchannels combined.

\subsubsection{Temporal precision}

For successfully matched segments, temporal precision was evaluated by comparing the estimated onset and offset boundaries with those of the corresponding human reference segment. Absolute onset and offset errors were defined as

\begin{equation}
    \Delta t_{\mathrm{on},i}
    =
    \left|
    t_{\mathrm{ref,on},i}
    -
    t_{\mathrm{est,on},i}
    \right|,
    \label{eq:onset_error}
\end{equation}

and

\begin{equation}
    \Delta t_{\mathrm{off},i}
    =
    \left|
    t_{\mathrm{ref,off},i}
    -
    t_{\mathrm{est,off},i}
    \right|,
    \label{eq:offset_error}
\end{equation}

where $t_{\mathrm{ref}}$ and $t_{\mathrm{est}}$ denote the reference and estimated boundaries, respectively. 

\begin{quote}
\textit{Example: if a human reference segment spans 2.0--3.0\,s and the matched pipeline segment spans 2.1--2.9\,s, the onset and offset errors are both 0.1\,s.}
\end{quote}

Temporal precision was summarized from the pooled absolute boundary errors of successfully matched segments across all conversations, separately for turn onsets, turn offsets, backchannel onsets, and backchannel offsets. The median absolute error (MedAE) and interquartile range (IQR) were used to describe the typical error and its spread. Mean absolute error (MAE) and standard deviation (SD) were additionally reported to characterize the influence of larger errors. For comparisons between listening conditions, MedAE was calculated separately for each conversation to provide one temporal-error measure per conversation.

\section{Results}

\subsection{Effect of context-based post-processing}

\begin{figure*}[htbp]
    \centering
    \includegraphics[width=\textwidth]
    {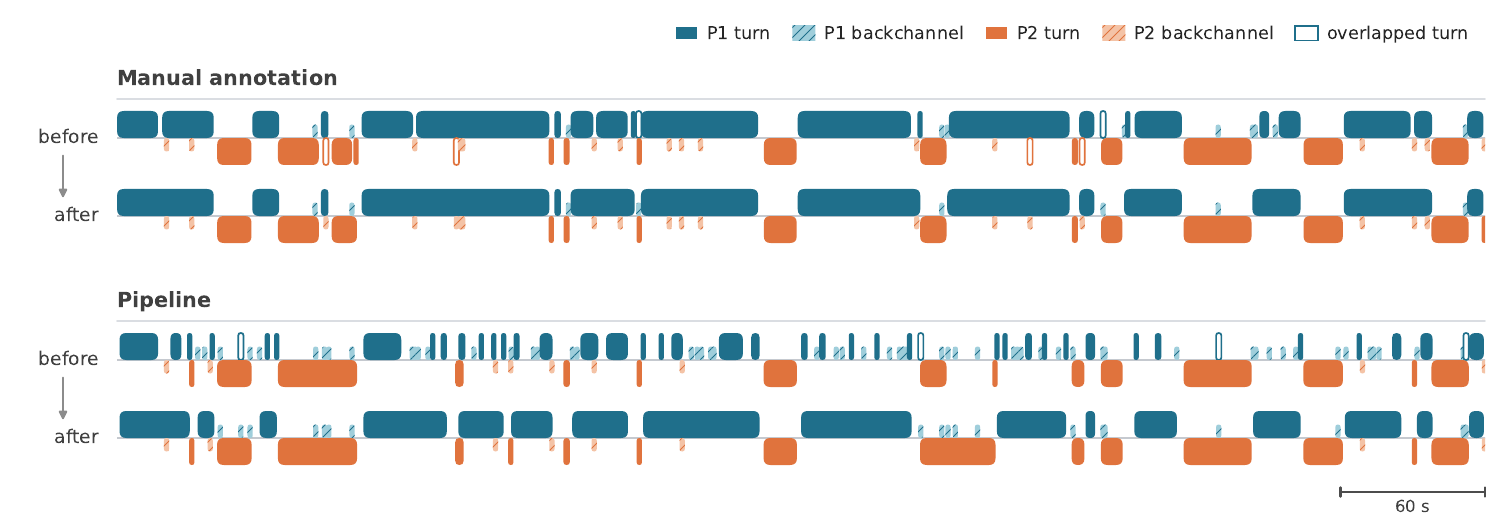}
    \caption{Example showing the effect of context-based post-processing on one conversation. Manual reference annotations (top) and pipeline output (bottom) are shown before and after post-processing. Colours distinguish turns and backchannels produced by the two participants, while outlined segments indicate overlapped turns.}
    \label{fig:merge_example}
\end{figure*}

Figure~\ref{fig:merge_example} illustrates the effect of context-based post-processing (Stage VII) on one example conversation. Before post-processing, the pipeline output is considerably more fragmented than the manual annotation. After post-processing, many of these fragments are joined into longer same-speaker units, and the broader pattern of speaker alternation is visually similar to that of the manual annotation. Agreement for individual backchannels was less consistent, with some manually annotated backchannels not captured by the pipeline.

For this example, post-processing reduced the manual annotation from 69 to 55 segments and the pipeline output from 125 to 58 segments (Table~\ref{tab:merge_example}). Across all 99 conversations, the mean number of segments per conversation decreased from 186.6 to 172.2 for the manual annotations (7.7\%) and from 158.8 to 154.1 for the pipeline (3.0\%). The corresponding mean number of turns decreased from 78.8 to 71.8 for the manual annotations and from 51.4 to 49.3 for the pipeline, while the mean number of backchannels decreased from 107.8 to 100.4 and from 107.5 to 104.8, respectively. 

Together, the example illustrates the intended use of the pipeline as an assistive first pass. It provides a structured approximation of the conversation that preserves the broader turn-taking pattern, while leaving individual events available for subsequent human review. This study did not directly evaluate annotation time, and potential annotation efficiency gains are therefore not quantified here.

\begin{table}[htbp]
\centering
\resizebox{\columnwidth}{!}{%
\begin{tabular}{llrrrr}
\toprule
\textbf{Source} & \textbf{Stage} &
\textbf{Turn} & \textbf{BC} &
\textbf{Overlapped} & \textbf{Total} \\
\midrule
Manual annotation & Before & 36 & 27 & 6 & 69 \\
                  & After  & 28 & 27 & 0 & 55 \\
\addlinespace
Pipeline          & Before & 55 & 66 & 4 & 125 \\
                  & After  & 31 & 27 & 0 & 58 \\
\bottomrule
\end{tabular}%
}
\caption{Segment counts for the conversation shown in
Fig.~\ref{fig:merge_example}, before and after context-based post-processing
of the manual reference annotation and pipeline output.}
\label{tab:merge_example}
\end{table}

\subsection{Overall pipeline performance}

Across the full set of 99 conversations, pipeline-generated turns and
backchannels were evaluated against the human reference annotations using the
matching criteria described in Section~\ref{sec:eval}. Mean detection reliability was similar for turns ($F_1 = 0.624 \pm 0.143$) and backchannels ($F_1 = 0.618 \pm 0.160$), with an overall $F_1$ score of $0.621 \pm 0.151$ across conversations (Table~\ref{tab:global_performance}).

For successfully matched segments, absolute boundary errors were smaller for
backchannels than for turns. Median absolute errors were 0.130\,s for
backchannel onsets and 0.180\,s for backchannel offsets, compared with
0.150\,s and 0.160\,s for turn onsets and offsets, respectively. Mean
absolute errors showed a larger difference, with values of 0.994\,s and
0.984\,s for turn onsets and offsets, compared with 0.152\,s and 0.219\,s
for backchannels. These absolute errors should be interpreted in relation to
segment duration, as backchannels are typically substantially shorter than
turns. Nevertheless, the large difference between MedAE and MAE within the
turn category indicates that most matched turn boundaries had relatively
small errors, while a smaller number of large errors increased the mean.

\begin{table}[htbp]
\centering
\resizebox{\columnwidth}{!}{%
\begin{tabular}{lccc}
\toprule
\textbf{Metric} & \textbf{Mean $F_1$} & \textbf{MedAE (s)} & \textbf{MAE (s)} \\
\midrule

\multicolumn{4}{l}{\textbf{Detection reliability}} \\
\quad Overall       & $0.621 \pm 0.151$ & -- & -- \\
\quad Turn          & $0.624 \pm 0.143$ & -- & -- \\
\quad Backchannel   & $0.618 \pm 0.160$ & -- & -- \\

\midrule
\multicolumn{4}{l}{\textbf{Temporal precision}} \\
\quad Turn onset          & -- & $0.150$ [0.100--0.910] & $0.994 \pm 2.383$ \\
\quad Turn offset         & -- & $0.160$ [0.080--0.530] & $0.984 \pm 2.662$ \\
\quad Backchannel onset   & -- & $0.130$ [0.090--0.160] & $0.152 \pm 0.185$ \\
\quad Backchannel offset  & -- & $0.180$ [0.120--0.230] & $0.219 \pm 0.260$ \\

\bottomrule
\end{tabular}%
}
\caption{Global pipeline performance against human reference annotations across 99 conversations from 33 dyads. Detection reliability is reported as the mean $F_1$ score $\pm$ standard deviation (SD) across conversations. Temporal precision pools the absolute boundary error of every correctly matched segment ($n=19,958$) and summarises that one distribution twice: as the median with its inter-quartile range (MedAE) and as the mean $\pm$ SD (MAE). Absolute errors are bounded below by zero and right-skewed, so the median is paired with an IQR rather than an SD. All temporal errors are in seconds.}
\label{tab:global_performance}
\end{table}

\subsection{Performance across listening conditions}

\begin{figure*}[t]
  \centering
  \includegraphics[width=0.96\textwidth]{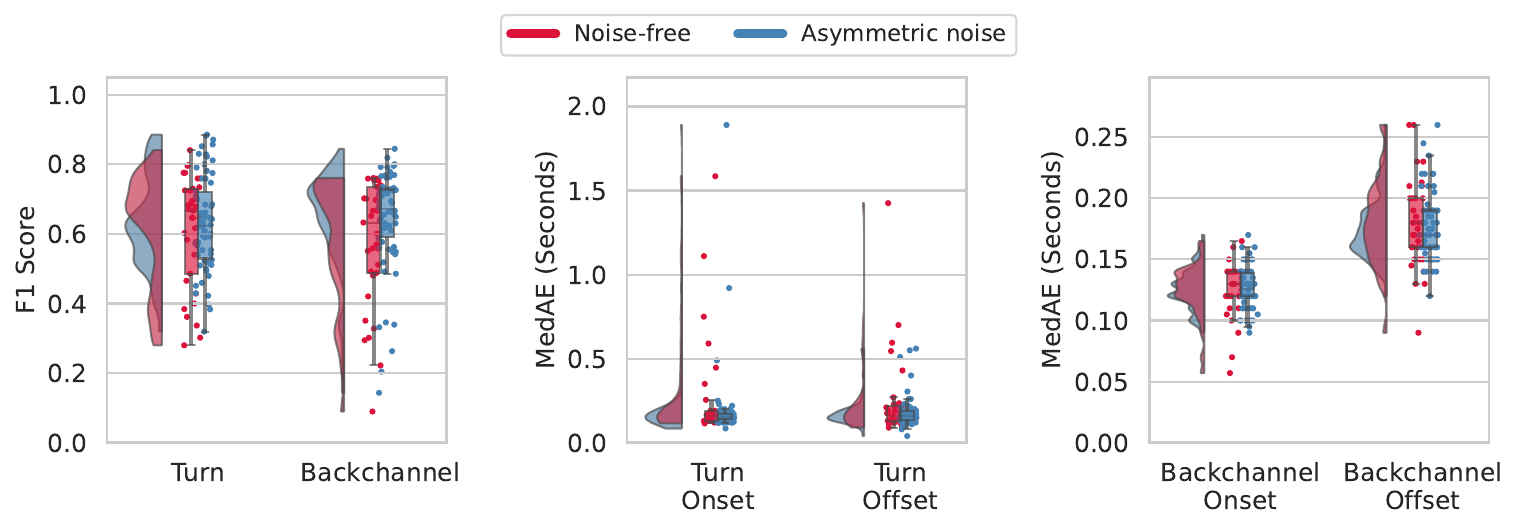}
  \caption{Pipeline performance across noise-free and asymmetric-noise listening
conditions for all 99 conversations. \textbf{Left}: $F_1$ scores for turn and
backchannel detection. \textbf{Centre}: median absolute error (MedAE) for turn
onset and offset boundaries. \textbf{Right}: MedAE for backchannel onset and
offset boundaries. Differences between listening conditions were assessed
using two-sided Mann--Whitney $U$ tests; none of the comparisons was
statistically significant.}
  \label{fig:results_panel_all}
\end{figure*}

To examine whether pipeline performance differed between listening conditions,
we compared all 99 conversations using two-sided Mann--Whitney $U$ tests (Fig.~\ref{fig:results_panel_all}). Detection reliability did not differ significantly between noise-free and asymmetric-noise conditions for either turn $F_1$ ($p=.885$) or backchannel $F_1$ ($p=.113$). Temporal precision likewise did not differ significantly for turn onset ($p=.389$), turn offset ($p=.747$), backchannel onset ($p=.809$), or backchannel offset ($p=.627$). Hence, there was no evidence of a systematic difference in pipeline performance between the two listening conditions.

\subsection{Case study: Benchmarking against inter-annotator agreement}

To examine sensitivity to parameter selection and benchmark pipeline performance against human inter-annotator agreement, we conducted a case study using four conversations independently annotated by two human annotators. The same segment-matching criteria used for pipeline--human evaluation were applied to the human annotations, providing a directly comparable measure of segment-level agreement.

We evaluated 80 parameter combinations, comprising 10 values of the rVAD threshold ($\tau_{\mathrm{VAD}}$; 0.1--1.0) and eight values of the entropy threshold ($\tau_H$; 0.5--3.0). Applied to each of the four conversations, this resulted in 320 pipeline runs.

Across the 320 runs, mean pipeline detection reliability was $F_1=0.626 \pm 0.225$ overall, $0.668 \pm 0.179$ for turns, and $0.585 \pm 0.256$ for backchannels (Table~\ref{tab:case_study}). Human inter-annotator agreement on the same four  conversations was higher, with $F_1=0.845 \pm 0.118$ overall, $0.868 \pm 0.051$ for turns, and $0.821 \pm 0.168$ for backchannels.

Temporal differences were also smaller between human annotators. Median
absolute boundary errors for the pipeline were 0.190\,s and 0.210\,s for
turn onsets and offsets, respectively, compared with 0.080\,s and 0.110\,s
between human annotators. For backchannels, the corresponding values were
0.140\,s and 0.200\,s for the pipeline and 0.050\,s and 0.040\,s between
human annotators.

\begin{figure*}[t]
  \centering
    \includegraphics[width=0.96\textwidth]{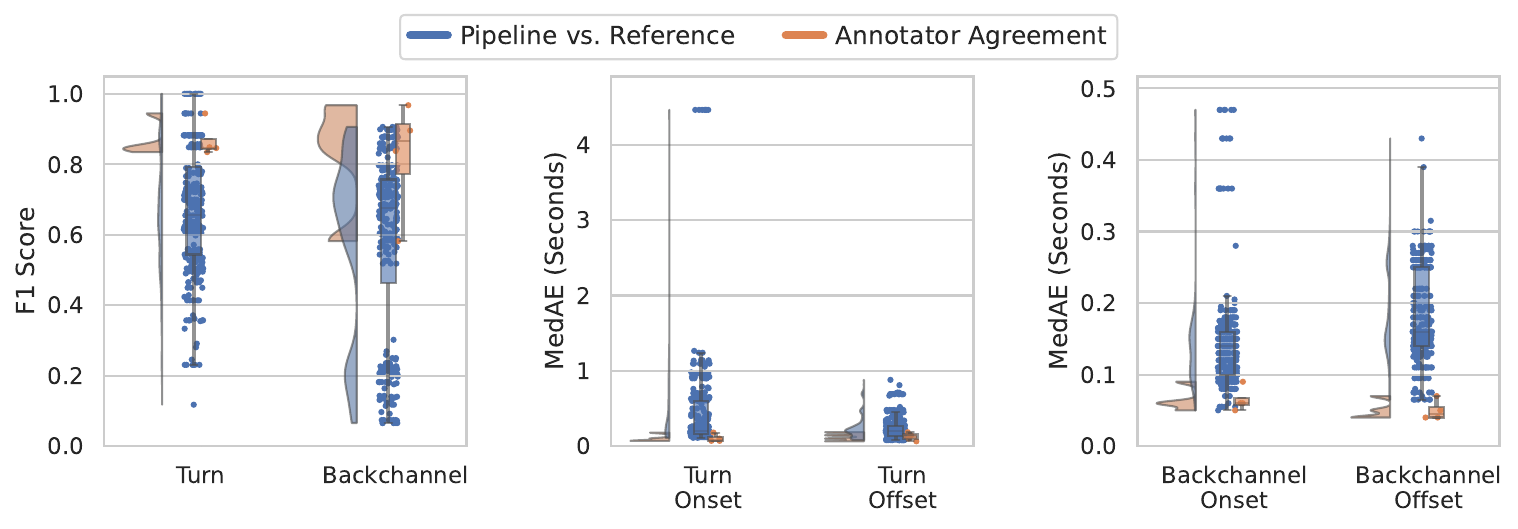}
    \caption{Case-study comparison of pipeline--human agreement across 320 pipeline runs (80 parameter combinations applied to four conversations) and human inter-annotator agreement on the same four conversations. \textbf{Left}: $F_1$ scores for turn and backchannel detection. \textbf{Centre}: median absolute error (MedAE) for turn onset and offset boundaries. \textbf{Right}: MedAE for backchannel onset and offset boundaries. Pipeline distributions reflect variation across parameter combinations, whereas human agreement reflects the four independently annotated conversations. The distributions are shown descriptively and were not subjected to inferential group comparison because the repeated pipeline runs are not independent observations.}
    \label{fig:results_panel:case}
\end{figure*}

As shown in Fig.~\ref{fig:results_panel:case}, pipeline performance also varied substantially across parameter combinations. The pipeline distributions were much broader than the inter-annotator distributions, particularly for backchannel $F_1$ and temporal boundary errors, and included a greater number of large-error cases. One conversation produced a turn $F_1$ of 1.0 for some parameter settings. This conversation contained long stretches of speech from one participant with few interruptions or short responses from the interlocutor, resulting in a comparatively simple turn structure. Overall, performance varied substantially across parameter settings and across the four conversations included in the case study.

No inferential statistical comparison was performed between pipeline-human and human–human agreement. The 320 pipeline observations represent repeated parameter combinations applied to the same four conversations and are therefore not independent observations, while the human benchmark is based on only those four conversations. Inferential tests treating these values as independent samples would therefore overstate the effective sample size.

\begin{table}[htbp]
  \centering
  \resizebox{\columnwidth}{!}{%
  \begin{tabular}{lcc}
    \toprule
    \textbf{Metric} & \textbf{Pipeline} & \textbf{Human agreement} \\
    & (320 runs) & (4 conversations) \\
    \midrule
    \multicolumn{3}{l}{\textbf{Detection reliability ($F_1$, mean $\pm$ SD across runs)}} \\
    \quad Overall      & $0.626 \pm 0.225$ & $0.845 \pm 0.118$ \\
    \quad Turn                 & $0.668 \pm 0.179$ & $0.868 \pm 0.051$ \\
    \quad Backchannel          & $0.585 \pm 0.256$ & $0.821 \pm 0.168$ \\
    \midrule
    \multicolumn{3}{l}{\textbf{Temporal precision --- MedAE (median [IQR], s)}} \\
    \quad Turn onset           & $0.190$ [0.100--1.600] & $0.080$ [0.050--0.202] \\
    \quad Turn offset          & $0.210$ [0.090--0.930] & $0.110$ [0.060--0.289] \\
    \quad Backchannel onset    & $0.140$ [0.090--0.180] & $0.050$ [0.040--0.080] \\
    \quad Backchannel offset   & $0.200$ [0.130--0.260] & $0.040$ [0.020--0.070] \\
    \midrule
    \multicolumn{3}{l}{\textbf{Temporal precision --- MAE (mean $\pm$ SD, s)}} \\
    \quad Turn onset           & $1.688 \pm 3.907$ & $0.649 \pm 2.213$ \\
    \quad Turn offset          & $1.044 \pm 2.236$ & $0.561 \pm 1.207$ \\
    \quad Backchannel onset    & $0.168 \pm 0.257$ & $0.083 \pm 0.145$ \\
    \quad Backchannel offset   & $0.234 \pm 0.289$ & $0.096 \pm 0.204$ \\
    \bottomrule
  \end{tabular}%
  }
  \caption{Case study: pipeline performance across 320 grid runs
  (80 parameter combinations on 4 conversations) benchmarked against human
  inter-annotator agreement on the same conversations. $F_1$ is the mean
  $\pm$ SD across runs. Temporal precision pools the absolute boundary error
  of every correctly matched segment ($n=54,146$ pipeline, $n=928$ human)
  and summarises that distribution twice: as the median with its
  inter-quartile range (MedAE) and as the mean $\pm$ SD (MAE). The pipeline
  pool spans the whole grid, so each conversation contributes its segments
  once per parameter combination.}
  \label{tab:case_study}
\end{table}

\section{Discussion}

This study evaluated an automated pipeline for identifying turns and backchannels in spontaneous dyadic conversation, intended for use within a semi-automated annotation workflow. A central challenge in large-scale conversation research is not only identifying conversational units, but doing so consistently across recordings. Manual annotation requires repeated judgements about turn boundaries and interactional function, which can be time-consuming and may vary between annotators. An automated first pass offers a way to apply the same operational criteria systematically across a corpus, with subsequent human review to resolve ambiguous cases. This can support more standardized and reproducible annotation while retaining human judgement where interactional context is important.

Across 99 Danish conversations, the pipeline achieved similar segment-level
detection reliability for turns and backchannels, with an overall $F_1$ score
of approximately 0.62. For successfully matched segments, median temporal
errors were relatively small, with median absolute errors of 0.150\,s and
0.160\,s for turn onsets and offsets, respectively. Mean absolute errors were
substantially larger, at approximately 1\,s for both boundaries. The mean
therefore does not represent the typical boundary difference between matched
pipeline and human segments. Instead, the distribution contains a smaller
number of substantially larger errors. Such errors may arise in part from the
temporal complexity of spontaneous speech, including hesitations, filled
pauses, and disfluencies \cite{shriberg2001errrr}, but they also identify
cases in which the segmentation procedure requires further improvement.
Backchannel boundaries showed smaller differences between median and mean
errors, indicating fewer extreme timing errors.

Pipeline performance did not differ significantly between the noise-free and asymmetric-noise listening conditions. Importantly, the experimental manipulation altered the listening context under which participants produced speech rather than introducing noise into the recorded microphone signal. These results therefore show that the pipeline performance was comparable across the recordings obtained from the two experimental listening conditions.

The four-conversation case study provided a complementary comparison with human inter-annotator agreement. Pipeline-human agreement was numerically lower and more variable than human-human agreement for both detection and temporal precision. These comparisons were descriptive rather than inferential. The 320 pipeline runs represent repeated parameter combinations applied to the same four conversations and cannot be treated as independent
observations, while the human comparison was based on only four conversations. The case study should therefore be interpreted as a descriptive benchmark rather than as a statistical comparison of pipeline and human performance.

The case study also highlights an important issue in evaluating automated conversation annotation: human reference annotations themselves contain variability. The two annotators differed in both segment classification and temporal boundary placement. Classifying turns and backchannels can require judgement about conversational context and communicative function \cite{carletta1996assessing,artstein2008survey}. Human agreement should therefore not be treated as a strict performance ceiling but a descriptive benchmark. A larger, systematic study of inter-annotator agreement is needed to characterize variability in both conversational-unit classification and temporal boundary estimation.

Performance also varied across pipeline parameter settings, particularly for backchannel detection. This indicates sensitivity to the rVAD and entropy thresholds and reinforces the importance of parameter selection. Backchannels present a particular challenge because they are often short and may contain little lexical information. A further limitation is the reliance on automatic transcription for conversational-unit classification. Short or non-lexical backchannels may be omitted or inaccurately transcribed by the ASR system, which can affect the lexical-entropy measure and consequently the assigned turn or backchannel label. Transcription accuracy was not evaluated separately in the present study. Future work should evaluate ASR performance specifically for these short conversational responses and evaluate parameter selection on independent data.

Two additional limitations concern the practical use and generalizability of the pipeline. Annotation time was not measured directly, so the extent to which the proposed workflow reduces manual annotation effort remains to be quantified. The present evaluation was also limited to dyadic, separate-channel Danish conversation. Future work should examine annotation efficiency and test the approach across languages, recording configurations, and interactional settings, including multi-party conversation.

\section{Conclusion}
This study presents an automated pipeline for identifying turns and backchannels in spontaneous dyadic conversation, designed for use within a semi-automated annotation workflow. Across 99 conversations, the pipeline provided a consistent first-pass segmentation with comparable detection reliability for turns and backchannels and relatively small median temporal errors for successfully matched segments. Although performance remained more variable than human inter-annotator agreement and was sensitive to parameter selection, the pipeline provides a consistent starting point for human review. By applying the same operational criteria across recordings, the pipeline can support more standardized and reproducible annotation procedures.

\section{Data and Code Availability}
The conversation recordings cannot be made publicly available due to
privacy restrictions. The code supporting the findings of this study is
available in the \href{https://github.com/haraldsr/Speech_VAD_Diarization_Transcription/tree/main}{project repository on GitHub}.

\section{Ethics}
All study procedures were approved by the Capital Region Committees of the Danish Science-Ethics Committee (reference H-16036391). All participants provided informed consent prior to participation.

\section{Author Contributions}
H.H.: Conceptualization, Methodology, Software, Formal analysis, Investigation, Data curation, Writing – original draft, Writing – review \& editing. H.V.S.R.: Conceptualization, Methodology, Software, Formal analysis, Writing – review \& editing. I.O.: Methodology, Writing – review \& editing. I.K.: Funding acquisition, Supervision, Writing – review \& editing.

\section{Funding}
This work is supported by the Carlsberg Foundation's Semper Ardens: Accelerate grant, ``Self, other, and we: mechanisms and dynamics of self-other integration in social observation and interaction'', no. CF22-1251, and the Innovation Fund Denmark grant, no. 4354-00015B.

\section{Declaration of competing interest}
The authors declare no competing interests.

\section{Generative AI Use Disclosure}
During the preparation of this manuscript, the authors used generative AI to help edit and refine the text and to assist with code. After using this tool, the authors reviewed and edited all content as needed and take full responsibility for it.

\bibliographystyle{IEEEtran}
\bibliography{references}

\end{document}